\documentclass[letterpaper]{article} 
\usepackage[preprint]{aaai2027}  
\usepackage[hyphens]{url}  
\usepackage{graphicx} 
\usepackage{natbib}  
\usepackage{caption} 
\usepackage{algorithm}
\usepackage{algorithmic}
\usepackage{amsmath}
\usepackage{amssymb}
\usepackage{newfloat}
\usepackage{listings}
\DeclareCaptionStyle{ruled}{labelfont=normalfont,labelsep=colon,strut=off} 
\floatstyle{ruled}
\newfloat{listing}{tb}{lst}{}
\floatname{listing}{Listing}
\usepackage{pifont}
\usepackage{multirow}
\usepackage{booktabs}

\title{ForgeDrive: Bidirectional Cross-Conditioning for Unified Visual-Action Generation in Autonomous Driving}
\author{
    Xuchang Zhong\textsuperscript{\rm 1\dag}, He Zheng\textsuperscript{\rm 1}, Chenxu Zhao\textsuperscript{\rm 1}, Tianxiong Lv\textsuperscript{\rm 1}\corresponding\ddag, Hangqi Fan\textsuperscript{\rm 1}\\
    Bohua Wang\textsuperscript{\rm 1}, Yushan Liu\textsuperscript{\rm 1,2\dag}, Li Gao\textsuperscript{\rm 1}, Zhihao Liao\textsuperscript{\rm 1}, Leigang Luo\textsuperscript{\rm 1}, Congyang Zhao\textsuperscript{\rm 1}, Yang Cai\textsuperscript{\rm 1}\\
}
\affiliations{
    \textsuperscript{\rm 1}Amap, Alibaba Group
    \textsuperscript{\rm 2}Tsinghua University
}

\begin{document}

\maketitle

\renewcommand\thefootnote{}\footnotetext{\dag Work done during the internship at Amap, Alibaba Group.}
\renewcommand\thefootnote{}\footnotetext{\ddag Project Leader.}
\begin{abstract}
World-model-based autonomous driving endows the model with the ability to understand scene evolution. Yet this promise is undermined by the prevailing imagine-then-act paradigm, which allows errors from the more challenging visual generation stage to cascade into action planning. We introduce ForgeDrive, a unified autoregressive diffusion framework with visual-action cross-conditioning that closes this gap through act-then-imagine paradigm. ForgeDrive factorizes the future as a sequence of per-timestep frame-action pairs, intertwining each action with its corresponding visual observation. During training, we decouple the diffusion timesteps of the two modalities and introduce a UniDiffuser-style noise scheduler to get the ability to infer either modality from its counterpart and deepen understanding of relationships between images and actions. At inference, we propose a novel act-then-imagine inference paradigm, and find that at each step, action generation is a capability internalized via co-training with the world model, requiring no clean future frame as a prerequisite at inference time; instead, the generated action can improve the accuracy of future frame generation, which in turn enhances the quality of the next action. Additionally, we augment each step with future ego-status prediction, further sharpening planning ability. Extensive experiments on NAVSIM demonstrate that ForgeDrive not only unifies driving simulation, planning, and visual odometry into a single model, but also outperforms existing strong planners without any post-training strategy.
\end{abstract}

\begin{figure*}[t]
\centering
\includegraphics[width=0.9\textwidth]{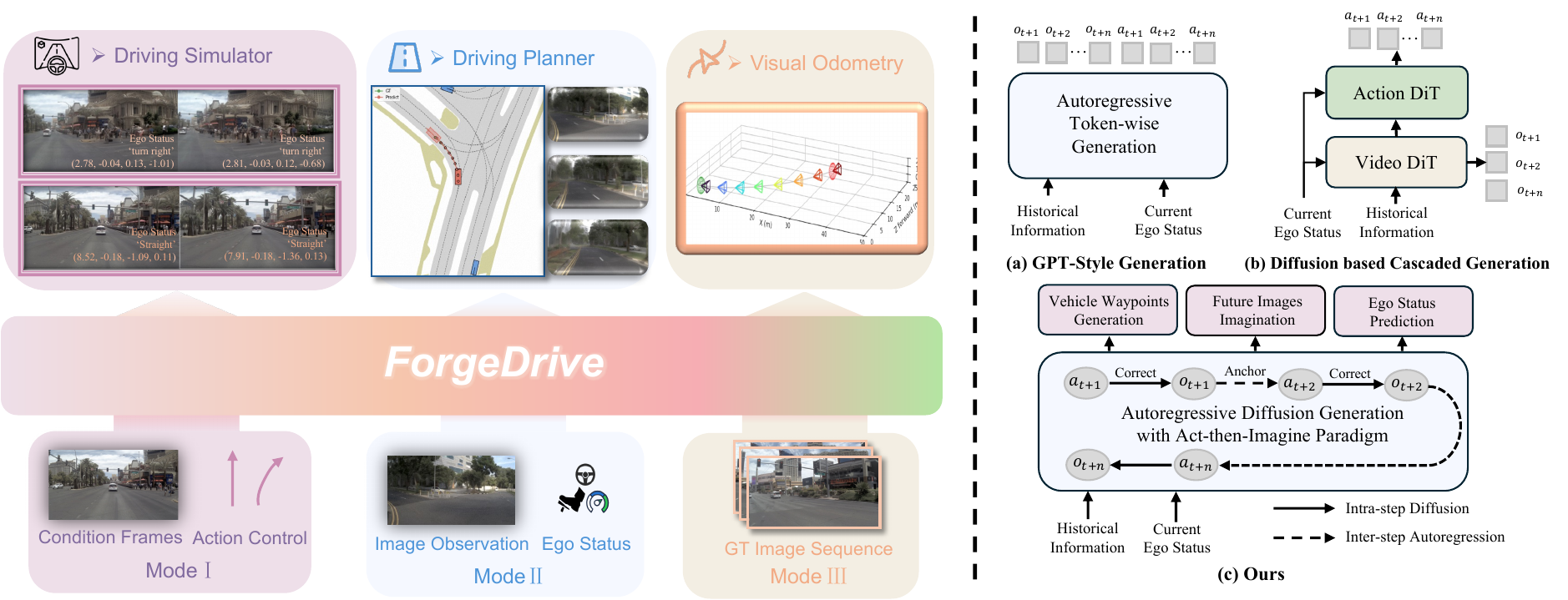} 
\caption{ForgeDrive: a unified autoregressive diffusion framework for autonomous driving. \textbf{Left:} ForgeDrive unifies three capabilities---driving simulator, driving planner, and visual odometry---within a single model. \textbf{Right:} Comparison of different world-model paradigms. Our model creates a virtuous cycle that progressively tightens both modalities over the horizon: within step, the predicted action corrects the corresponding image generation; across steps, the generated image anchors the next action.}
\label{fig1_intro}
\end{figure*}

\section{Introduction}
\label{sec:intro}
A central aspiration of autonomous driving is to endow agents with the ability to anticipate the future before acting. World models fulfill this need by learning to predict future states directly from raw video, internalizing environment dynamics to foresee the consequences of potential actions without costly annotation. This combination of predictive capability and data efficiency has made world models increasingly central to autonomous driving.

The key to world-model-based approaches \cite{zhang2026future, yang2026resim, li2026sgdrive,zheng2025world4drive} lies in enabling the model to faithfully capture how the scene evolves, thereby yielding more robust planning decisions. Existing approaches that jointly generate images and actions can be broadly classified into two paradigms. GPT-style methods ~\cite{chen2025drivinggpt, jia2023adriver, zhao2026forecasting, zeng2026futuresightdrive, li2025drivevla}, as shown in Figure \ref{fig1_intro} (a), tokenize both future observations and actions, and model them jointly as a unified token sequence via next-token prediction. However, the interaction between the two modalities is mediated solely by causal attention over a flat token sequence, providing only an indirect form of coupling that struggles to capture the fine-grained interplay between actions and scene evolution. Diffusion-based methods ~\cite{xia2026drivelaw}, as shown in Figure \ref{fig1_intro} (b), first synthesize future frames via a video generation DiT and then condition a downstream trajectory predictor on the resulting features, tightening the connection between visual predictions and planned actions. Yet unlike world action model in embodied manipulation \cite{ye2026world, li2026causal}, autonomous driving involves larger scene variations and longer prediction horizons, making future frame synthesis based solely on pixel-level motion extrapolation especially vulnerable to compounding errors that ultimately corrupt trajectory prediction. Furthermore, all such pipelines generate the entire video segment in one shot, leaving the planner no opportunity to rectify accumulated visual inaccuracies. A natural question arises: \textit{how can we forge a tighter, bidirectional coupling between future visual states and actions, so that each modality continuously anchors and corrects the other?}

To address this question, we argue that such bidirectional coupling presupposes the ability to infer either modality from its counterpart: recovering actions from visual observations and synthesizing future frames from actions. To this end, we propose \textbf{ForgeDrive}, an autoregressive diffusion framework that unifies visual and action generation through cross-conditioning. Rather than committing to all frames in one shot, ForgeDrive builds the future progressively as a sequence of frame--action pairs, thereby endowing the model with the ability to use actions to correct future visual predictions. To equip the model with the prerequisite bidirectional inference capability, we decouple the diffusion timesteps of the two modalities and introduce a UniDiffuser-style noise scheduler \cite{bao2023one} that enables mixed training across three complementary objectives: inverse dynamics modeling, action-conditioned video generation, and joint video-action prediction. During inference, we observe that action prediction is comparatively easy and largely mastered through co-training with the world model, requiring only a coarse sense of the upcoming state at inference, whereas future frame synthesis is substantially harder and benefits significantly from action guidance. This asymmetry motivates an \textit{act-then-imagine} strategy: at each timestep the action is resolved first to correct the subsequent imagination, and the resulting faithful visual context in turn forces the next action to be better informed, creating a virtuous cycle that progressively tightens both modalities over the horizon. Beyond frame--action generation, we further augment each timestep with a prediction of the next-step ego status. This auxiliary objective compels the model to maintain physical consistency between the predicted action and the resulting vehicle dynamics, providing a complementary supervision signal that sharpens trajectory quality.

Owing to the above architectural and strategic designs, ForgeDrive also unifies multiple capabilities within a single model. (1) \textbf{Full-state driving simulator}: it can generate future frames and the corresponding ego status conditioned on given observations and a specified action sequence. (2) \textbf{Driving planner}: it reasons about plausible future trajectory directly from the current observations and ego status. (3) \textbf{Visual odometry}: it recovers the relative pose between frames from an input image sequence. This unified capability set eliminates the need for separate, task-specific models and allows each function to benefit from the representations learned by the others.

In summary, our contributions are as follows:

\begin{itemize}
\item We propose \textbf{ForgeDrive}, a unified autoregressive diffusion framework that predicts future frames, actions, and ego status for autonomous driving, where multi-objective mixed training deepens cross-modal understanding and unifies driving simulation, planning, and visual odometry within a single model.
\item We investigate the causal relationship between future state and action generation, revealing that the counter-intuitive \textit{act-then-imagine} order yields more robust predictions than the prevailing imagine-then-act paradigm.
\item Extensive experiments on NAVSIM benchmark demonstrate ForgeDrive achieves strong planning performance under purely supervised learning, without reinforcement learning or trajectory scoring, outperforming several competitive existing planners.
\end{itemize}

\begin{figure*}[ht]
\centering
\includegraphics[width=0.9\textwidth]{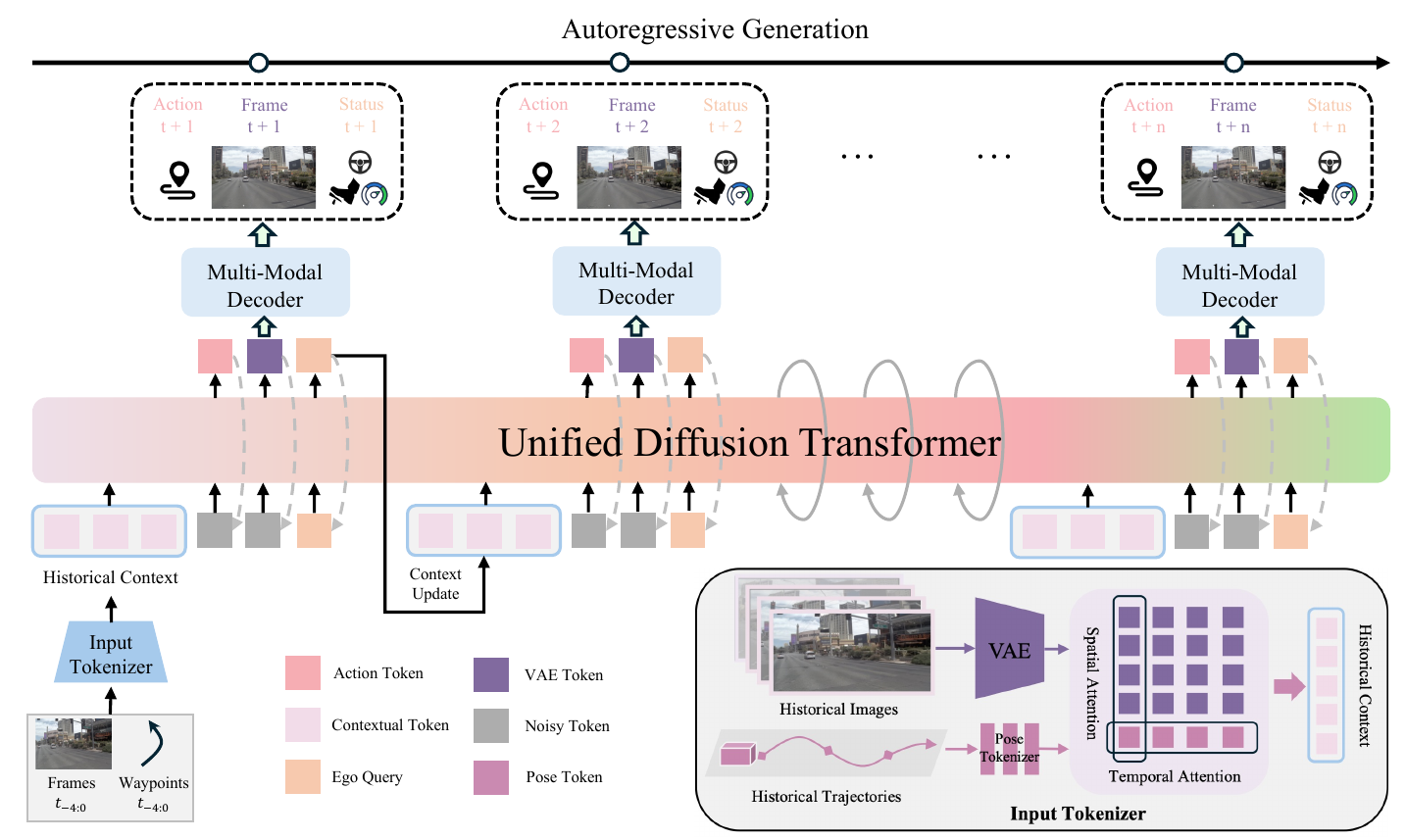} 
\caption{Overall architecture of ForgeDrive. At each autoregressive step, a sliding window of $T$ historical frames with poses and the current ego state is encoded into a context feature via a frozen autoencoder and spatial-temporal attention. The \textbf{UniDiT} then denoises image and action tokens under decoupled diffusion timesteps, with full cross-modal attention throughout, and outputs the next frame, action, and ego state in one forward pass. The prediction is fed back into the window to update context for the next step, rolling out the full horizon autoregressively.}
\label{fig2_pipline}
\end{figure*}

\section{Related Works}

\subsection{End to End Autonomous Driving}
End-to-end autonomous driving \cite{jiang2024vadv2, jia2025drivetransformer, yin2026diffrefiner}aims to directly map raw sensor observations to planning decisions within a unified framework. These methods typically require explicit perception modules to extract intermediate representations. Query-based approaches such as UniAD~\cite{hu2023planning}, VAD~\cite{jiang2023vad}, and SparseDrive~\cite{sun2025sparsedrive} chain perception, prediction, and planning through learnable queries, producing deterministic trajectory outputs via regression. More recent works adopt generative formulations to capture the multimodal nature of driving: GenAD~\cite{zheng2024genad} treats trajectory prediction as a latent generative process, DiffusionDrive~\cite{liao2025diffusiondrive} employs a truncated diffusion model for diverse trajectory sampling, and GoalFlow~\cite{xing2025goalflow} leverages flow matching to generate goal-conditioned trajectories. However, all these methods operate solely on the action modality without modeling future visual observations, forgoing the opportunity to leverage predicted scene context for more informed planning.

\subsection{World Models for Autonomous Driving}
In autonomous driving, world models endow the model with the ability to predict the future, thereby improving the  quality of trajectories. Representative works such as GAIA-1~\cite{hu2023gaia}, DriveDreamer~\cite{wang2024drivedreamer}, Drive-WM~\cite{wang2024driving}, and Vista~\cite{gao2024vista} focus on generating high-quality future frames, leveraging autoregressive transformers or video diffusion models. Recent works bridge video and action generation within a single model: DrivingGPT~\cite{chen2025drivinggpt}, PWM ~\cite{zhao2026forecasting} and ADriver-I~\cite{jia2023adriver} jointly predict frames and actions via next-token prediction; Epona~\cite{zhang2025epona} uses twin parallel DiTs from a shared causal latent; DriveLAW~\cite{xia2026drivelaw} unifies both in a shared latent space by serially connecting video and action DiTs. However, these methods either treat the two modalities as a flat token sequence or generate them sequentially without exploiting the interplay between them. ForgeDrive instead decouples the diffusion timesteps of image and action tokens, creating bidirectional learning signals through cross-conditioning, and reverses the conventional generation order via an act-then-imagine inference strategy.

\section{Methods}

\subsection{Overview}
Given a sequence of $T$ historical camera observations $\{\mathbf{I}_{-T+1}, \dots, \mathbf{I}_0\}$ along with the corresponding ego-vehicle poses and the current ego status $\mathbf{s}_0$, the goal is to predict a sequence of $H$ future frame--action--status tuples $\{(\hat{\mathbf{I}}_1, \hat{\mathbf{a}}_1, \hat{\mathbf{s}}_1), \dots, (\hat{\mathbf{I}}_H, \hat{\mathbf{a}}_H, \hat{\mathbf{s}}_H)\}$. As illustrated in Figure ~\ref{fig2_pipline}, ForgeDrive operates autoregressively along the temporal axis: at each future timestep, the model takes a sliding window of the most recent $T$ frames along with their associated poses and the current ego status, predicts the next frame--action--status tuple via a Unified Diffusion Transformer, and appends the prediction to the window for the subsequent step.

\subsection{Model Architecture}
\label{sec:MA}

\paragraph{Input tokenization.}
Each raw image $\mathbf{I}_t$ is encoded into spatial tokens $\mathbf{z}_t \in \mathbb{R}^{L \times C}$ by a frozen pre-trained DCAE encoder \cite{chen2025deep}, where $L = H' \times W'$. The inter-frame ego-vehicle pose $(\Delta x_t, \Delta y_t, \Delta \psi_t)$ is mapped to pose tokens $\mathbf{e}_t^{\text{pose}} \in \mathbb{R}^{L_{\text{pose}} \times D}$ via Fourier feature embedding and learned projectors. The two groups are concatenated per frame to form $\mathbf{h}_t = [\,\mathbf{e}_t^{\text{pose}};\; \mathbf{z}_t\,] \in \mathbb{R}^{L_{\text{total}} \times D}$. The $T$-frame sequence, augmented with learnable temporal embeddings, is processed by  spatial-temporal blocks. Each block alternates between causal temporal attention and full spatial attention:
\begin{gather}
    \tilde{\mathbf{h}}_t^{(l)} = \mathrm{TemporalAttn}\!\bigl(\mathbf{h}_t^{(l)},\; \{\mathbf{h}_{\tau}^{(l)}\}_{\tau \leq t}\bigr), \\
    \mathbf{h}_t^{(l+1)} = \mathrm{SpatialAttn}\!\bigl(\tilde{\mathbf{h}}_t^{(l)}\bigr),
\end{gather}
The output of the last condition frame serves as the historical context feature $\mathbf{c} = \mathbf{h}_0 \in \mathbb{R}^{L_{\text{total}} \times D}$. Additionally, the current ego state $\mathbf{s}_t$ is encoded into a single token $\mathbf{e}^{\text{ego}} \in \mathbb{R}^{1 \times D}$ by summing the outputs of an MLP applied to the dynamics and a learned embedding of the navigation command.

\paragraph{Unified Diffusion Transformer.}
Existing two-stage pipelines employ separate networks for video generation and trajectory prediction, preventing information flow between the two tasks during denoising. ForgeDrive instead unifies both modalities into a single Diffusion Transformer (DiT).

The UniDiT operates on the concatenation of four token groups, each augmented with a learned type embedding:
\begin{equation}
    \mathbf{X} = \mathbf{c} \oplus \mathbf{z}_{t_\text{img}}^{\text{img}} \oplus \mathbf{z}_{t_\text{act}}^{\text{act}} \oplus e^{\text{ego}}\;\in \mathbb{R}^{(L_{c} + L_{\text{img}} + L_{\text{act}} + 1) \times D},
\end{equation}
where $\oplus$ denotes concatenation along the token dimension, $\mathbf{z}_{t_\text{img}}^{\text{img}}$, $\mathbf{z}_{t_\text{act}}^{\text{act}}$ are the noisy image and action tokens at their respective diffusion timesteps.

Following the MMDiT architecture~\cite{esser2024scaling, labs2025flux}, the sequence $\mathbf{X}$ is processed through $N_d$ double-stream blocks and $N_s$ single-stream blocks. All blocks perform joint attention over the full token sequence, enabling information exchange across modalities:
\begin{gather}
    \mathbf{Q},\, \mathbf{K},\, \mathbf{V} = \mathrm{Proj}^m_{q,k,v}(\mathbf{X}^{(l)}), \\
    \mathbf{A} = \mathrm{Attn}(\mathbf{Q}, \mathbf{K}, \mathbf{V}),
\end{gather}
where $m \in \{\text{img}, \text{act}, \text{cond}\}$ indexes the modality, and $\mathrm{Proj}^m$ denotes per-modality projections in double-stream blocks (shared in single-stream blocks). Each modality stream is independently modulated via AdaLN \cite{peebles2023scalable} conditioned on its corresponding timestep embedding, allowing the network to adapt its processing to the noise level of each modality separately. The UniDiT thus learns to estimate the joint velocity field while simultaneously predicting the next-step ego status from the ego token:
\begin{equation}
    (\hat{\mathbf{v}}_{\text{img}},\, \hat{\mathbf{v}}_{\text{act}},\, \hat{\mathbf{e}}_{t+1}^{ego}) = \mathrm{UniDiT}_\theta\!\bigl(\mathbf{z}_{t_\text{img}}^{\text{img}},\, \mathbf{z}_{t_\text{act}}^{\text{act}}, e^{ego}_t \mid \mathbf{c} \big)
\end{equation}

\paragraph{Decoding and Output.}
The decoding process of UniDiT follows a flow matching formulation, iteratively denoising the image and trajectory latent from pure Gaussian noise to the target distribution. The denoised image latent is then decoded by a DCAE decoder to produce the final frame, while the trajectory is mapped to the action space through an MLP head. After the final denoising step, the ego dynamics and driving command are decoded from the ego token's hidden representation:
\begin{equation}
    \hat{\mathbf{s}}_{\text{dyn}}, \hat{\mathbf{s}}_{\text{cmd}} = D_{\text{dyn}}(\mathbf{e}^{\text{ego}}), \; D_{\text{cmd}}(\mathbf{e}^{\text{ego}}),
\end{equation}
where $D_{\text{dyn}}$, $D_{\text{cmd}}$ are MLP heads. $\hat{\mathbf{s}}_{\text{dyn}}\in \mathbb{R}^{4}$ represents ego velocity and acceleration prediction, and $\hat{\mathbf{s}}_{\text{cmd}} \in \mathbb{R}^{4}$ is logits over navigation commands. For autoregressive generation, rather than feeding back these decoded outputs, we directly recycle the latent representations: the denoised image latent is appended to the context window as updated observation, the ego token is forwarded as the current ego status, and the predicted action is accumulated into the global trajectory.

\begin{figure}[htbp]
\centering
\includegraphics[width=\linewidth]{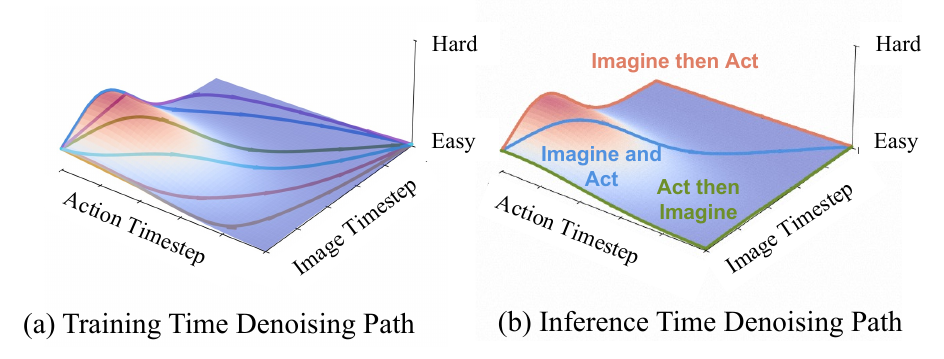} 
\caption{Denoising trajectories during training and inference. Training: we independently vary the diffusion timesteps of the two modalities to cover diverse denoising conditions. Inference: we explore three denoising schedules with different modality resolution orders.}
\label{fig3_denoise}
\end{figure}

\subsection{Training and Inference Schedule}
\label{sec:schedule}

Bidirectional coupling between visual predictions and actions requires the model to not only denoise both modalities jointly, but also to recover each modality conditioned on the other, thereby enabling one modality to refine and correct the other. This capability is established at training time and operationalized at inference time through a unified design we term cross-conditioning. The training phase addresses the question of \textit{how to equip the model with the ability to infer either modality from its counterpart}; the inference phase addresses the complementary question of \textit{how to best leverage this ability to achieve robust mutual correction during autoregressive rollout}.

\paragraph{Training.}
To equip the model with the ability to infer either modality from its counterpart, we design a training scheme that exposes the model to diverse cross-modal generation conditions within a single unified framework. Inspired by the multi-modal diffusion framework of UniDiffuser~\cite{bao2023one}, as shown in Figure \ref{fig3_denoise} (a), we decouple the diffusion timesteps of the two modalities and sample $t_{\text{img}}$ and $t_{\text{act}}$ independently from $\mathcal{U}(0, 1)$. Under this scheme, the model learns a unified conditional distribution:
\begin{equation}
    p_\theta(\mathbf{z}^{\text{img}},\, \mathbf{z}^{\text{act}} \mid \mathbf{c},\, t_{\text{img}},\, t_{\text{act}}),
\end{equation}
which subsumes three complementary learning objectives as special cases depending on the sampled timestep pair:
\begin{itemize}
    \item \textbf{Inverse dynamics modeling} ($t_{\text{img}} \!\to\! 1,\; t_{\text{act}} \!\to\! 0$): the image tokens are nearly clean while the action tokens are heavily noised, reducing to $p_\theta(\mathbf{z}^{\text{act}} \mid \mathbf{z}^{\text{img}}, \mathbf{c})$ and training the model to recover actions from visual observations.
    \item \textbf{Action-conditioned video generation} ($t_{\text{img}} \!\to\! 0,\; t_{\text{act}} \!\to\! 1$): the action tokens are nearly clean while the image tokens are heavily noised, reducing to $p_\theta(\mathbf{z}^{\text{img}} \mid \mathbf{z}^{\text{act}}, \mathbf{c})$ and training the model to synthesize future frames guided by the intended action.
    \item \textbf{Joint video--action prediction} ($t_{\text{img}} \!\approx\! t_{\text{act}}$): both modalities carry comparable noise, training the model to denoise them simultaneously.
\end{itemize}

By default, the model is trained with teacher forcing, where ground-truth serve as context for subsequent autoregressive steps. However, at inference time the model must condition on its own imperfect predictions, creating a discrepancy that can lead to compounding drift. To bridge this gap, we adopt scheduled sampling: with a probability that warms up linearly during training, the model's own predictions replace the ground-truth values in the context window, gradually acclimating the model to autoregressive self-conditioning.

\begin{table*}[ht]
    \centering
    \setlength{\tabcolsep}{3.8pt}
    \begin{tabular*}{\linewidth}{@{\extracolsep{\fill}}lccccccccccc}
        \toprule
        \textbf{Method} 
            & \textbf{NC}$\uparrow$ & \textbf{DAC}$\uparrow$ & \textbf{DDC}$\uparrow$ & \textbf{TL}$\uparrow$ & \textbf{EP}$\uparrow$ 
            & \textbf{TTC}$\uparrow$ & \textbf{LK}$\uparrow$ & \textbf{HC}$\uparrow$ & \textbf{EC}$\uparrow$ & \textbf{EPDMS}$^*$$\uparrow$ & \textbf{EPDMS}$\uparrow$ \\
        \midrule
        TransFuser \cite{chitta2022transfuser}& 96.9 & 89.9 & 97.8 & 99.7 & 87.1 & 95.4 & 92.7 & 98.3 & 87.2 & 76.7 & $-$ \\
        Hydra-MDP++ \cite{li2025hydra++} & 97.2 & 97.5 & 99.4 & 99.6 & 83.1 & 96.5 & 94.4 & 98.2 & 70.9 & 81.4 & $-$ \\
        DriveSuprim \cite{yao2026drivesuprim} & 97.5 & 96.5 & 99.4 & 99.6 & 88.4 & 96.6 & 95.5 & 98.3 & 77.0 & 83.1 & $-$ \\
        ARTEMIS \cite{feng2025artemis} & 98.3 & 95.1 & 98.6 & 99.8 & 81.5 & 97.4 & 96.5 & 98.3 & $-$  & 83.1 & $-$ \\
        DiffusionDriveV2 \cite{zou2025diffusiondrivev2}& 97.7 & 96.6 & 99.2 & 99.8 & \textbf{88.9} & 97.2 & 96.0 & 97.8 & \textbf{91.0} & 85.5 & 87.5 \\
        DriveVLA-W0 \cite{li2025drivevla} & 98.5 & \textbf{99.1} & 98.0 & 99.7 & 86.4 & 98.1 & 93.2 & 97.9 & 58.9 & 86.1  & $-$\\
        Recogdrive \cite{li2025recogdrive} & 98.3 & 95.2 & 98.3 & 99.8 & 87.1 & 97.5 & 96.6 & \textbf{99.5} & 86.5 & 83.6  & $-$ \\
        Epona \cite{zhang2025epona} &97.1 &95.7 &99.3 &99.7 &88.6 &96.3 &97.0 &98.0 &67.8 &$-$ &85.1 \\
        DreamAD \cite{yang2026dreamerad} &98.0 &97.2 &99.5 &99.8 &87.8 &97.4 &97.5 &98.3 &72.4 & $-$ & 87.7 \\
        Latent-WAM \cite{wang2026latent} & 98.1 & 97.3 & \textbf{99.6} & 99.8 & 87.7 & 97.3 & 97.6 & 98.1 & 87.3 & $-$  & 89.3 \\
        DriveFuture \cite{hong2026drivefuture} &98.8 &\textbf{99.1} &\textbf{99.6} &\textbf{99.9} &86.6 & \textbf{98.4} &96.4 &98.3 &74.8 &86.4 &89.9 \\
        \midrule
        \textbf{Ours} & \textbf{98.9} & 97.6 & \textbf{99.6} & 99.8 & 87.2 & \textbf{98.4} & \textbf{98.1} & 98.4 & 86.5 & \textbf{86.6} & \textbf{90.3} \\
        \bottomrule
    \end{tabular*}
    \caption{Comparison with state-of-the-art methods on the NavSim v2 benchmark. \textbf{Bold} indicates the best result in each column.}
    \label{tab:main_results_v2}
\end{table*}

\begin{table*}[ht]
    \centering
    \setlength{\tabcolsep}{4pt}
    \small
    \begin{tabular}{l|cccccc|c}
        \toprule
        \textbf{Metric} & \textbf{Epona} & \textbf{WoVoGen} & \textbf{Vista} & \textbf{DrivingGPT} & \textbf{Uni-World VLA} & \textbf{PWM} & \textbf{Ours} \\
        \midrule
        FVD $\downarrow$       & 82.8       & 417.7      & 89.4       & 142.6     & 141.8     & 85.95 & \textbf{69.2} \\
        Max Duration / FPS     & 120s / 2Hz & 2.5s / 2Hz & 15s / 10Hz & 4s / 2Hz  & 4s / 2Hz  & 4s / 2hz & 120s / 2Hz \\
        Dataset                & nuScenes   & nuScenes   & nuScenes   & NavSim    & NavSim    & NavSim & NavSim \\
        View                   & Front      & Multi      & Front      & Front     & Front     & Front  & Front \\
        \bottomrule
    \end{tabular}
    \caption{Quantitative results of video generation. Methods: Epona~\cite{zhang2025epona}, WoVoGen~\cite{lu2024wovogen}, Vista~\cite{gao2024vista}, DrivingGPT~\cite{chen2025drivinggpt}, PWM~\cite{zhao2026forecasting}, Uni-World VLA~\cite{liu2026uni}.}
    \label{tab:metric_video}
\end{table*}


\paragraph{Inference.}
The training scheme above equips ForgeDrive with the ability to generate either modality conditioned on the other. The inference phase addresses the complementary question of how to best operationalize this ability for robust autoregressive rollout. Because the decoupled timesteps allow the model to follow arbitrary denoising paths, as shown in Figure~\ref{fig3_denoise}~(b), we identify three canonical generation orders, each prescribing a distinct conditioning hierarchy between actions and images:
\begin{itemize}
    \item \textbf{Imagine-then-Act} ($t_{\text{img}}$: $0{\to}1$, then $t_{\text{act}}$: $0{\to}1$): the image tokens are denoised first while action tokens remain pure noise; the resolved future frame is then frozen as a condition for action denoising.
    \item \textbf{Imagine-and-Act} ($t_{\text{img}} {=} t_{\text{act}}$: $0{\to}1$): both modalities are denoised jointly with a shared timestep schedule.
    \item \textbf{Act-then-Imagine} ($t_{\text{act}}$: $0{\to}1$, then $t_{\text{img}}$: $0{\to}1$): the action tokens are denoised first while image tokens remain pure noise; the resolved action is then frozen as a condition for image denoising.
\end{itemize}
We provide a detailed discussion on the distinct behaviors of these three modes in Table~\ref{tab:ablation_infer} and adopt the \textit{act-then-imagine} schedule as our final inference strategy.

\subsection{Training Objective} 
ForgeDrive is trained with rectified flow ~\cite{liu2022flow}. The standard velocity-prediction losses are applied to the image and action tokens, denoted as $\mathcal{L}_{\text{img}}$ and $\mathcal{L}_{\text{act}}$, respectively. Additionally, we supervise the predicted next-step ego status with a combined loss that regresses the dynamics $(v_x, v_y, a_x, a_y)$ via MSE and classifies the navigation command $\mathbf{d}$ via cross-entropy:
\begin{equation}
    \mathcal{L}_{\text{ego}} = \|\hat{\mathbf{s}}_{\text{dyn}} - \mathbf{s}_{\text{dyn}}\|^2 + \mathrm{CE}(\hat{\mathbf{s}}_{\text{cmd}},\, \mathbf{s}_{\text{cmd}}).
\end{equation}
The total loss is:
\begin{equation}
    \mathcal{L} = \mathcal{L}_{\text{img}} + \lambda_{\text{act}}\,\mathcal{L}_{\text{act}} + \lambda_{\text{ego}}\,\mathcal{L}_{\text{ego}}.
\end{equation}
where $\lambda_{\text{act}}$ and $\lambda_{\text{ego}}$ are scalar weights balancing the action and ego-state losses against the image loss.

\begin{table*}[ht]
    \centering
    \setlength{\tabcolsep}{4pt}
    \begin{tabular}{lccccccc}
        \toprule
        \textbf{Methods} 
            & \textbf{Input}
            & \textbf{NC}$\uparrow$ 
            & \textbf{DAC}$\uparrow$ 
            & \textbf{TTC}$\uparrow$ 
            & \textbf{Comf}$\uparrow$ 
            & \textbf{EP}$\uparrow$ 
            & \textbf{PDMS}$\uparrow$ \\
        \midrule
        \multicolumn{8}{l}{\textit{Traditional End-to-End Methods}} \\
        Uniad \cite{hu2023planning} & C & 97.8 & 91.9 & 92.9 & \textbf{100} & 78.8 & 83.4 \\
        TransFuser \cite{chitta2022transfuser} & C\&L & 97.7 & 92.8 & 92.8 & \textbf{100} & 79.2 & 84.0 \\
        Hydra-MDP \cite{li2024hydra} & C\&L & 98.3 & 96.0 & 94.6 & \textbf{100} & 78.7 & 86.5 \\
        DiffusionDrive \cite{liao2025diffusiondrive} & C\&L & 98.2 & 96.2 & 94.7 & \textbf{100} & 82.2 & 88.1 \\
        \midrule
        \multicolumn{8}{l}{\textit{World Model Methods}} \\
        Driving GPT \cite{chen2025drivinggpt} & SC & 98.9 & 90.7 & 94.9 & 95.6 & 79.7 & 82.4 \\
        FSdrive \cite{zeng2026futuresightdrive} & SC & 98.2 & 93.8 & 93.3 & 99.9 & 80.1 & 85.1 \\
        Epona \cite{zhang2025epona} & SC & 97.9 & 95.1 & 93.8 & 99.9 & 80.4 & 86.2 \\
        PWM \cite{zhao2026forecasting} & SC & 98.6 & 95.9 & 95.4 & \textbf{100} & 81.8 & 88.1 \\
        DriveVLA-w0 \cite{li2025drivevla} & SC & 98.7 & 96.2 & 95.1 & \textbf{100} & 82.2 & 88.4 \\
        DriveLaW \cite{xia2026drivelaw} & SC & \textbf{99.0} & 97.1 & \textbf{96.7} & \textbf{100} & 81.3 & 89.1 \\
        Uni-World VLA \cite{liu2026uni} & SC & 98.7 & 96.7 & 96.1 & \textbf{100} & 83.2 & 89.4 \\
        \midrule
        \textbf{Ours} & SC & 98.9 & \textbf{97.6} & 96.3 & \textbf{100} & \textbf{83.8} & \textbf{90.2} \\
        \bottomrule
    \end{tabular}
    \caption{
        Comparison with state-of-the-art methods on the NavSim v1 benchmark.
        \textbf{Bold} indicates the best result in each column.
    }
    \label{tab:main_results_v1}
\end{table*}


\begin{table}[t]
    \centering
    \small
    \setlength{\tabcolsep}{4pt}
    \begin{tabular}{l|c|cc}
        \toprule
        \textbf{Method} & \textbf{ADE $\downarrow$} & \textbf{Dyn.\ MAE $\downarrow$} & \textbf{Cmd Acc $\uparrow$} \\
        \midrule
        DPVO       & 0.09 & -- & -- \\
        \textbf{Ours}   & 0.11 & 0.43/0.03/0.31/0.21 & 94.7\% \\
        \bottomrule
    \end{tabular}
    \caption{Trajectory and ego status prediction results. Dynamic MAE is reported in the order of ($v_x, v_y, a_x, a_y$).}
    \label{tab:metrics_ego}
\end{table}

\section{Experiment}

\subsection{Experimental Setup}
\paragraph{Dataset and Metrics.} We evaluate our method on the NAVSIM benchmark \cite{dauner2024navsim, cao2025pseudo}, a high-fidelity simulation dataset derived from nuPlan \cite{caesar2021nuplan}. To comprehensively validate the planning robustness of our approach, we report results across three official splits: NAVSIM v1 navtest, NAVSIM v2 navtest/navhard. We adopt the PDMS for v1 and the EPDMS for v2 and navhard as our primary evaluation metrics. To evaluate video generation quality, we utilize Fréchet Video Distance (FVD) \cite{unterthiner2018towards} to assess the statistical realism of predicted frames. Furthermore, we quantify ego status prediction performance using two distinct criteria: Mean Absolute Error (MAE) for ego dynamics and classification accuracy for driving commands. We also report Average Displacement Error (ADE) to measure trajectory accuracy under the inverse dynamics model.

\paragraph{Implementation Details.}
Our ForgeDrive with 3B parameters adopts a Mixture-of-Transformer \cite{liang2024mixture} architecture, with the video branch built on Epona \cite{zhang2025epona}, a FLUX-based \cite{labs2025flux} DiT. Training proceeds in two stages: the video branch is first adapted on nuPlan at 2 Hz for 30k steps, after which the newly integrated action branch is jointly trained on NAVSIM for 30 epochs. Training is conducted on 8$\times$ NVIDIA H20 GPUs, using a total batch size of 16 and an input resolution of 512$\times$1024 on 2 Hz sequences. The model is optimized with AdamW \cite{loshchilov2017decoupled} (learning rate $1\times10^{-4}$, weight decay $5\times10^{-2}$) and loss weights $\lambda_{\text{act}}=1.0$, $\lambda_{\text{ego}}=0.01$. At inference, trajectory evaluation applies 2 denoising steps per modality (4 in total), while image-quality evaluation uses 10 image denoising steps.

\subsection{Main Results}

\paragraph{Planning Capability.} 
Table~\ref{tab:main_results_v1} compares our method with state-of-the-art approaches on NAVSIM v1. Trained solely with supervised learning on a single camera input, without reinforcement learning or learned trajectory scoring, our method achieves a PDMS of 90.2, surpassing traditional end-to-end methods with LiDAR input. Furthermore, compared to world-model-based methods, our model surpasses DriveLaW using a cascaded DiT generation approach by 1.1 PDMS, and outperforms PWM and Uni-World VLA using GPT-style methods by 2.1 and 0.8 PDMS. Table~\ref{tab:main_results_v2} presents results on the NAVSIM v2, which adopts a more challenging evaluation metric. Our method achieves 90.3 EPDMS with no drop from the PDMS score, alongside an excellent lane keeping score of 98.1, demonstrating the high quality of our trajectories. Compared with DreamAD, which also uses Epona as the video generation backbone but employs reinforcement learning, our method surpasses it by 2.6 EPDMS. Compared with the recent strong method Latent-WAM and DriveFuture, we outperform them by 1.0 and 0.4 EPDMS.

\paragraph{Video Generation Capability.}
Table~\ref{tab:metric_video} evaluates video generation quality and long-horizon capability on NAVSIM test split over a 4-second future horizon. We first compare generation fidelity against VQ-VAE \cite{van2017neural} methods: our diffusion-based approach achieves an FVD of 69.2, outperforming Uni-World VLA, PWM and DrivingGPT by a large margin, confirming that continuous latent diffusion produces more realistic and temporally coherent futures than discrete tokenization. In terms of long-horizon generation, our model matches Epona in supporting 120-second rollouts, far exceeding other methods. 

\begin{table*}[ht]
    \centering
    \setlength{\tabcolsep}{4pt}
    \small
    \begin{tabular}{ccc|cccccccccc}
        \toprule
             \textbf{Decouple Timestep} & \textbf{Sched. Samp.} &  \textbf{Ego Prediction}
            & \textbf{NC}$\uparrow$ & \textbf{DAC}$\uparrow$ & \textbf{DDC}$\uparrow$ & \textbf{TL}$\uparrow$ & \textbf{EP}$\uparrow$
            & \textbf{TTC}$\uparrow$ & \textbf{LK}$\uparrow$ & \textbf{HC}$\uparrow$ & \textbf{EC}$\uparrow$ & \textbf{EPDMS}$\uparrow$ \\
        \midrule
        \ding{55} & \ding{55} & \ding{55} & 98.3 & 96.4 & \textbf{99.6} & \textbf{99.8} & 87.1 & 97.4 & 97.0 & 98.2 & 75.3 & 87.2 \\
        \ding{51} & \ding{55} & \ding{55} & 98.6 & 97.1 & 99.5 & \textbf{99.8} & 87.4 & 98.2 & 98.0 & 98.3 & 83.1 & 89.1 \\
        \ding{51} & \ding{51} & \ding{55} & 98.7 & 97.4 & \textbf{99.6} & \textbf{99.8} & \textbf{87.9} & 98.2 & \textbf{98.1} & 98.2 & 83.8 & 89.7 \\
        \ding{51} & \ding{51} & \ding{51} & \textbf{98.9} & \textbf{97.5} & 99.5 & \textbf{99.8} & 87.3 & \textbf{98.4} & 98.0 & \textbf{98.4} & \textbf{86.2} & \textbf{90.0} \\
        \bottomrule
    \end{tabular}
    \caption{Ablation study on key training strategies. All the results of the above models are obtained in the Imagine-and-Act mode.}
    \label{tab:ablation_train}
\end{table*}

\paragraph{Additional Capability.} 
As shown in Table~\ref{tab:metrics_ego}, the results demonstrate the model's excellent capability in predicting ego status. Furthermore, Under the IDM mode, the model recovers driving trajectories directly from GT images, attaining a Mean ADE of 0.11\,m, comparable to DPVO~\cite{teed2023deep}, a dedicated visual odometry method (0.09\,m). This demonstrates that our world model, without any VO-specific design, acquires competitive localization ability as a byproduct of joint action--image generation.

\begin{figure}[htbp]
\centering
\includegraphics[width=\linewidth]{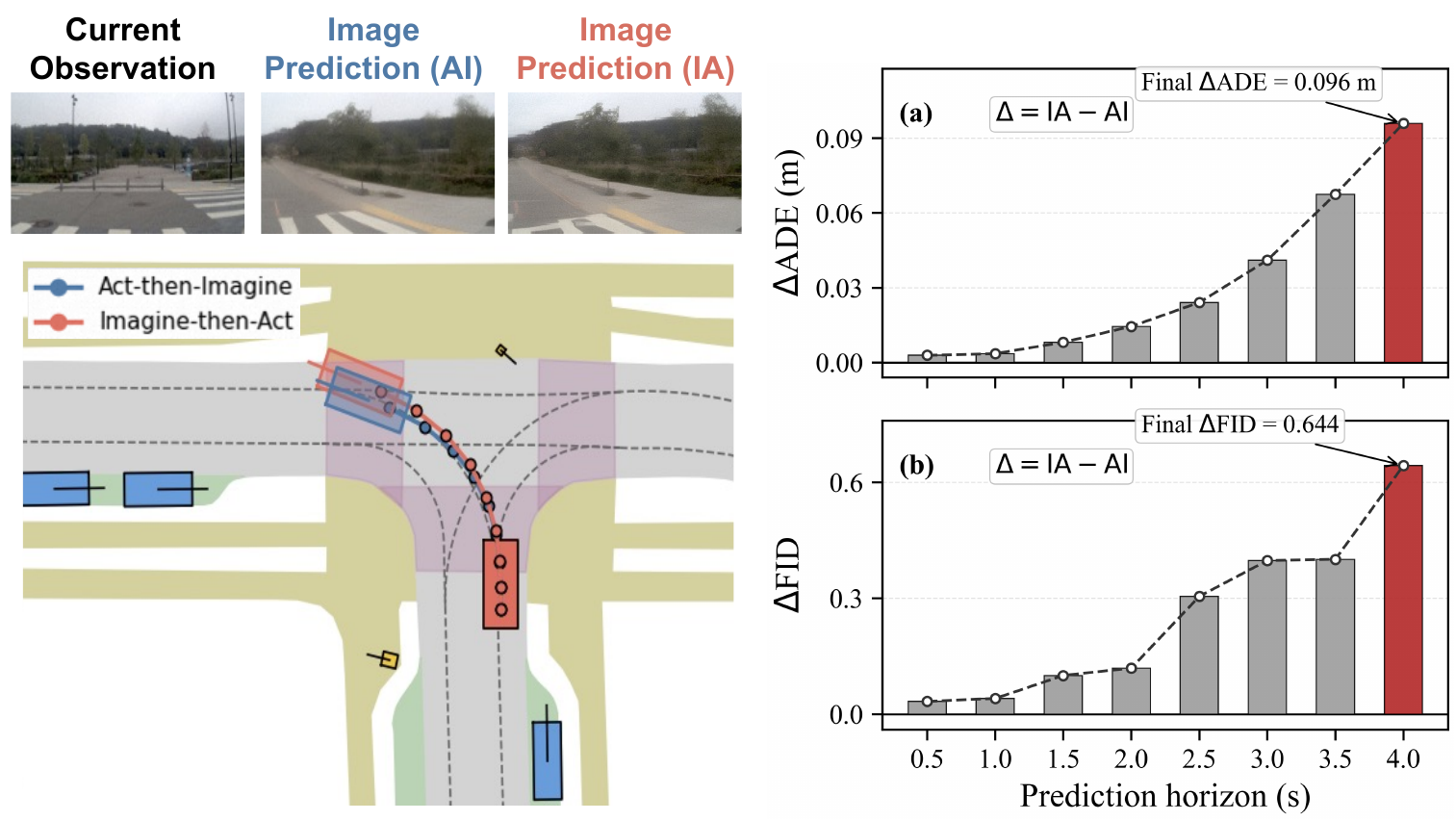} 
\caption{\textbf{Left:} Left: imagined future frames under different inference modes (IA: Imagine-then-Act, AI: Act-then-Imagine). Right: per-timestep trajectory ADE and frame FID, reported as differences between the two modes.}
\label{fig4_vis}
\end{figure}

\subsection{Ablation Study}

\paragraph{Effect of Training Strategy.} 
The ablation study in Table \ref{tab:ablation_train} validates the effectiveness of our training strategy. Introducing decoupled timestep scheduling yields substantial gains, with EPDMS improving from 87.2 to 89.1. This demonstrates that endowing the model with the ability to infer either modality from its counterpart can improve its understanding of the correspondence between image and action. Adding scheduled sampling increases EPDMS from 89.1 to 89.7, confirming that acclimating the model to its own predictions during training enhances robustness of the model. Furthermore, incorporating ego status prediction as an auxiliary objective provides additional improvements, raising EPDMS to 90.0 and increasing EC by 2.4 points. This shows that jointly training with self-motion supervision enhances the model’s scene understanding and decision-making consistency.

\paragraph{Effect of Inference Strategy.}
Table~\ref{tab:ablation_infer} reports planning performance under different inference orders. \textit{Act-then-Imagine} ranks first on both navhard and navtest, followed by \textit{Imagine-and-Act} and \textit{Imagine-then-Act}. On FVD, it also leads by a clear margin (69.2), indicating that action generation does not rely on clean future frames, whereas frame generation benefits from action guidance. Two probe settings further isolate the underlying mechanism. Under GT Future, where the model only generates actions at each step and future frames are directly taken from ground truth for the next step, the model reaches 92.8 EPDMS, showing that action generation is already internalized through co-training with the world model and requires no clean frames at inference. Conversely, masking out the noisy image tokens during action denoising in \textit{Act-then-Imagine} drops performance below \textit{Imagine-then-Act}, implying that future frames inform action by conveying an evolving trend rather than a clear future. Notably, the margin of \textit{Act-then-Imagine} over \textit{Imagine-then-Act} enlarges from +0.4 EPDMS on navtest to +1.6/+2.6 on navhard, suggesting that harder scenes amplify drift in action-free imagination, which in turn yields larger gains from action correction. Hence, \textit{Act-then-Imagine} not only exploits future information when resolving actions, but also leverages actions to correct imagination, achieving superior planning performance.

\begin{table}[t]
    \centering
    \setlength{\tabcolsep}{3pt}
    \small
    \begin{tabular}{l|cc|c|c}
        \toprule
        \multirow{2}{*}{\textbf{Infer. Strategies}}
            & \multicolumn{2}{c|}{\textbf{navhard}}
            & \multicolumn{1}{c|}{\textbf{navtest}}
            & \multirow{2}{*}{\textbf{FVD}$\downarrow$} \\
        \cmidrule(lr){2-3} \cmidrule(lr){4-4}
            & \textbf{EPDMS}$_1$$\uparrow$ & \textbf{EPDMS}$_2$$\uparrow$ & \textbf{EPDMS}$\uparrow$ & \\
        \midrule
        Imagine-then-Act           & 81.4 & 39.7 & 89.9 & 73.3 \\
        Imagine-and-Act            & 82.8 & 41.2 & 90.0 & 73.4 \\
        Act-then-Imagine           & \textbf{83.0} & \textbf{42.3} & \textbf{90.3} & \textbf{69.2} \\
        Act-then-Imagine$^\dagger$ & 79.3 & 41.8 & 89.2 & - \\
        \midrule
        GT Future                  & -    & -    & 92.8    & - \\
        Human                      & -    & -    & 94.5    & - \\
        \bottomrule
    \end{tabular}
    \caption{Ablation study on inference strategies. $^\dagger$ denotes applying future image mask during trajectory denoising. GT Future means ground-truth future frames are provided.}
    \label{tab:ablation_infer}
\end{table}

\subsection{Analysis of Error Accumulation}
Figure~\ref{fig4_vis} (left) compares the imagined future frames under different inference modes: in Imagine-then-Act, accumulated drift in the imagined frames misleads subsequent action recovery, whereas in Act-then-Imagine the resolved action corrects the subsequent imagination. Figure~\ref{fig4_vis} (right) quantifies this effect on both sides of the cycle. On the trajectory side, the ADE gap widens monotonically over the horizon: Act-then-Imagine reduces the ADE from 1.89 m to 1.79 m (5.1\% relative reduction) at 4s. The same trend holds on the imagination side: the per-timestep FID gap grows steadily and reaches 0.64 at 4 s (6.7\% relative reduction), indicating that action correction also keeps the imagined frames closer to the ground-truth distribution. Together, these results confirm that Act-then-Imagine closes a virtuous cycle in which actions correct imaginations and faithful imaginations anchor subsequent actions, jointly reducing drift in both modalities.

\section{Conclusion}
We presented ForgeDrive, a unified autoregressive diffusion framework that bridges visual generation and action planning through bidirectional cross-conditioning. By decoupling the diffusion timesteps of image and action tokens during training, our model acquires the ability to infer either modality from its counterpart, unifying driving simulation, planning, and visual odometry within a single architecture. Crucially, we identified and validated an \textit{act-then-imagine} inference paradigm that reverses the conventional generation order, allowing future states to inform action generation while predicted actions simultaneously correct future visual imagination, thereby preserving action quality and mitigating the error cascades inherent in imagine-then-act approaches. Extensive experiments on NAVSIM demonstrate that ForgeDrive achieves strong planning performance without post-training.

\newpage
\bibliography{aaai2027}

\newpage
\appendix

\twocolumn[
\begin{center}
  {\LARGE \bfseries Supplementary Material \par}
  \vspace{1em}
\end{center}
]

\section{More Details about Our Metrics}
\label{sec:pdms}

Our evaluation on NAVSIM v1 adopts the PDM Score (PDMS). Under this protocol, the policy commits to a 4-second trajectory, which is unrolled open-loop at 10 Hz through an LQR controller with a kinematic bicycle model, while surrounding agents replay their recorded motion. From this non-reactive rollout, several subscores in $[0,1]$ are computed and aggregated as
\begin{equation}
    \mathrm{PDMS} = \mathrm{NC} \times \mathrm{DAC} \times \frac{5 \times (\mathrm{EP} + \mathrm{TTC}) + 2 \times \mathrm{C}}{12},
    \label{eq:pdms}
\end{equation}
where NC and DAC serve as multiplicative penalty factors, and EP, TTC, and C enter a weighted average.

\paragraph{No at-fault Collision (NC).} This factor drops to 0 upon any collision with another road participant (vehicle, pedestrian, or cyclist), and to 0.5 for collisions involving static objects. Incidents for which the ego vehicle is not responsible, such as being hit while stationary, are exempted.

\paragraph{Drivable Area Compliance (DAC).} This factor becomes 0 if the ego vehicle leaves the legal drivable region at any point during the rollout.

\paragraph{Ego Progress (EP).} Progress along the route centerline is normalized by a safe upper-bound estimate provided by the privileged PDM-Closed planner. The ratio is truncated to $[0,1]$, with non-positive progress treated as zero when the upper bound falls below 5 meters.

\paragraph{Time to Collision (TTC).} This score equals 1 by default and switches to 0 once the ego vehicle, extrapolated under constant velocity and heading, exhibits a time-to-collision below a safety threshold at any simulation step.

\paragraph{Comfort (C).} Acceleration and jerk magnitudes of the executed trajectory are compared against predefined comfort thresholds.

For evaluation under the two-stage pseudo-simulation protocol of NAVSIM v2, we adopt the Extended PDM Score (EPDMS), which augments PDMS with additional rule-compliance and comfort criteria:
\begin{equation}
\begin{split}
    \mathrm{EPDMS} =\;& \mathrm{NC} \cdot \mathrm{DAC} \cdot \mathrm{DDC} \cdot \mathrm{TLC} \\
    &\cdot \frac{5(\mathrm{EP}+\mathrm{TTC}) + 2(\mathrm{LK}+\mathrm{HC}+\mathrm{EC})}{16}.
\end{split}
\label{eq:epdms}
\end{equation}
Besides the NC, DAC, EP, and TTC terms inherited from PDMS, EPDMS introduces five new subscores—Driving Direction Compliance (DDC), Traffic Light Compliance (TLC), Lane Keeping (LK), History Comfort (HC), and Extended Comfort (EC).

\paragraph{Driving Direction Compliance (DDC).} This factor penalizes travel against the designated lane direction, i.e., entering lanes of opposing traffic outside intersection areas.

\paragraph{Traffic Light Compliance (TLC).} This factor penalizes the ego vehicle for proceeding into a signalized intersection when its controlling traffic light is not green.

\paragraph{Lane Keeping (LK).} This score rewards trajectories that stay close to the current lane centerline and penalizes both prolonged straddling of lane boundaries and indecisive, partially committed lane-change maneuvers.

\paragraph{History Comfort (HC).} Comfort is judged on a continuous motion profile: the planned trajectory is concatenated with the vehicle's recorded motion over the preceding 1.5 seconds, and the nuPlan comfort criterion is applied to this combined trajectory.

\paragraph{Extended Comfort (EC).} This score verifies that the planned motion evolves smoothly, without abrupt changes between adjacent time steps.

\section{Additional Ablation Studies}

\paragraph{Effect of Inference Sampling Steps.}
Table~\ref{tab:sample} studies the effect of inference sampling steps on planning performance. With only 2 steps, the model already achieves a strong EPDMS of 89.9, demonstrating that the image-action distribution has been well learned during training. Increasing to 4 steps brings the best overall EPDMS of 90.3, with notable gains in EC (86.5) and TTC (98.4), indicating that slightly more denoising steps improve collision avoidance and temporal consistency. Beyond 4 steps, the trajectory planning performance changes only marginally: at 10 steps, the EPDMS remains competitive at 90.2, while NC and TTC reach their peak; further increasing to 20 steps yields a slight EP improvement but leads to a drop in EC and EPDMS. 

\paragraph{Effect of Image Resolution.}
Table~\ref{tab:resolution} investigates the effect of input image resolution. Increasing the resolution from $256{\times}512$ to $512{\times}1024$ brings consistent improvements on most metrics, most notably raising the Ego Progress (EP) from 86.6 to 87.3 and the Extended Comfort (EC) from 81.7 to 84.2, which together lift the overall EPDMS from 89.3 to 89.9. The gains in collision-related metrics suggest that higher-resolution inputs provide finer visual details critical for perceiving obstacles and navigating complex scenarios. Meanwhile, the No at Fault Collisions (NC) and Time to Collision (TTC) metrics remain stable or slightly decrease, indicating that the base resolution is already sufficient for basic collision avoidance. These results confirm that a moderate increase in resolution is beneficial for planning robustness, while also justifying our choice of $512{\times}1024$ as the default setting.

\section{More Details about Results on Navhard Split}
\paragraph{Comparison with other methods.} Table~\ref{tab:navhard-v2} compares our method with state-of-the-art approaches on the NAVSIM v2 (navhard split) benchmark. Our method achieves the highest overall EPDMS score of 34.8, surpassing the second-best DriveLaW by a notable margin of 4.2 points. In Stage~1, we attain the best results in NC (97.8), DAC (94.9), LK (96.9), and HC (97.8), with the DAC score exceeding the next best method by 5.8 points, demonstrating a clear advantage in drivable area compliance. In Stage~2, our method continues to lead in TLC (98.7), LK (49.9), and HC (96.6), while remaining competitive on the remaining metrics. In contrast, DiffusionDrive exhibits a sharp drop in safety-related metrics in Stage~2, and MindDrive, despite strong Stage~2 NC and DAC, trails in overall EPDMS due to weaker Stage~1 performance. The consistent strength across both stages enables our method to set a new state-of-the-art on this challenging benchmark.

\paragraph{Comparison between different inference strategy.} As shown in Table \ref{tab:navhard_ablation}, Act-then-Imagine achieves the best overall EPDMS of 34.8, outperforming Imagine-and-Act (33.9) and Imagine-then-Act (32.4). The advantage is particularly pronounced in safety-critical metrics: on Stage 2, Act-then-Imagine leads in DAC (70.9), TTC (77.5), and NC (81.5), indicating more collision-free and smoother driving behaviors. This validates the effectiveness of our paradigm—by generating actions first, the model leverages future-state-informed representations to produce reliable trajectories, which in turn serve as corrective signals for subsequent visual imagination, yielding safer and more coherent long-horizon predictions.

\section{Statistical Significance Analysis}
We assess the significance of the differences between \textit{Act-then-Imagine} and \textit{Imagine-then-Act} on navhard using a
paired Wilcoxon signed-rank test over per-scene metric differences, complemented by bootstrap confidence intervals (10{,}000 resamples). The trajectory-side gain is significant: at the 4 s horizon, the ADE of \textit{Imagine-then-Act} exceeds that of \textit{Act-then-Imagine} with a 95\% CI of $[0.032, 0.159]$ m ($p = 0.011$), confirming that action correction reliably suppresses long-horizon drift. For EPDMS detailed sub-metrics in navhard benchmark: \textit{Act-then-Imagine} significantly improves all collision-related metrics, including NC ($p = 1.4 \times 10^{-4}$), DAC ($p = 1.7 \times 10^{-5}$), DDC ($p = 1.1 \times 10^{-14}$), TTC ($p = 4.5 \times 10^{-5}$), and LK ($p = 0.0067$), whereas \textit{Imagine-then-Act} attains significantly higher EP ($p = 8.6 \times 10^{-32}$). We attribute this asymmetry to the drift of action-free imagination: the drift+ frames exaggerate scene motion, biasing IA toward longer and more aggressive trajectories that harvest extra progress in simple scenes where safety is easily satisfied, yet become hazardous in complex ones. By correcting such erroneous imaginations with resolved actions, AI achieves substantially safer behavior. Per-scene scores of compared baselines are unavailable, so the analysis is restricted to our own ablation pairs.

\section{Additional Details about UniDiT}
Figure \ref{fig10_unidit} illustrates the detailed architecture of our UniDiT model. To better preserve the video generation capability, the main video generation backbone follows the architecture of the existing state-of-the-art work \cite{zhang2025epona,labs2025flux}. On top of this, we integrate the input and prediction of actions and future ego status. The overall architecture can be divided into a double-stream block and a single-stream block. In the double-stream block, each input modality is projected through an independent MLP to learn modality-specific features. In the single-stream block, a shared MLP is employed to achieve better cross-modal fusion. Notably, independent timestep modulation is applied between the image stream and the action stream, endowing the model with the ability to infer one modality from the other.

\section{Inference Efficiency}
Table~\ref{tab:latency} reports the inference latency of ForgeDrive for a full 4 s autoregressive rollout, measured on a single NVIDIA H20 GPU. With the default 2+2 denoising steps, the model completes a rollout in 0.49 s at 256$\times$512 and 0.65 s at 512$\times$1024 input resolution. At the default resolution, ForgeDrive predicts the 4 s future in 0.65 s---over 6$\times$ faster than real time---demonstrating practical efficiency for online deployment.

\begin{table}[ht]
    \centering
    \setlength{\tabcolsep}{6pt}
    \begin{tabular}{lcc}
        \toprule
        \textbf{Input Resolution} & \textbf{Latency (s)}$\downarrow$ & \textbf{Real-time Factor}$\uparrow$ \\
        \midrule
        256$\times$512          & 0.49 & 8.2$\times$ \\
        \textbf{512$\times$1024} & 0.65 & 6.2$\times$ \\
        \bottomrule
    \end{tabular}
    \caption{Inference latency of a full 4 s autoregressive rollout under
    different input resolutions. Real-time factor is computed as the
    horizon length divided by latency.}
    \label{tab:latency}
\end{table}

\section{More Qualitative Visualization}

\begin{figure*}[t]
\centering
\includegraphics[width=0.9\textwidth]{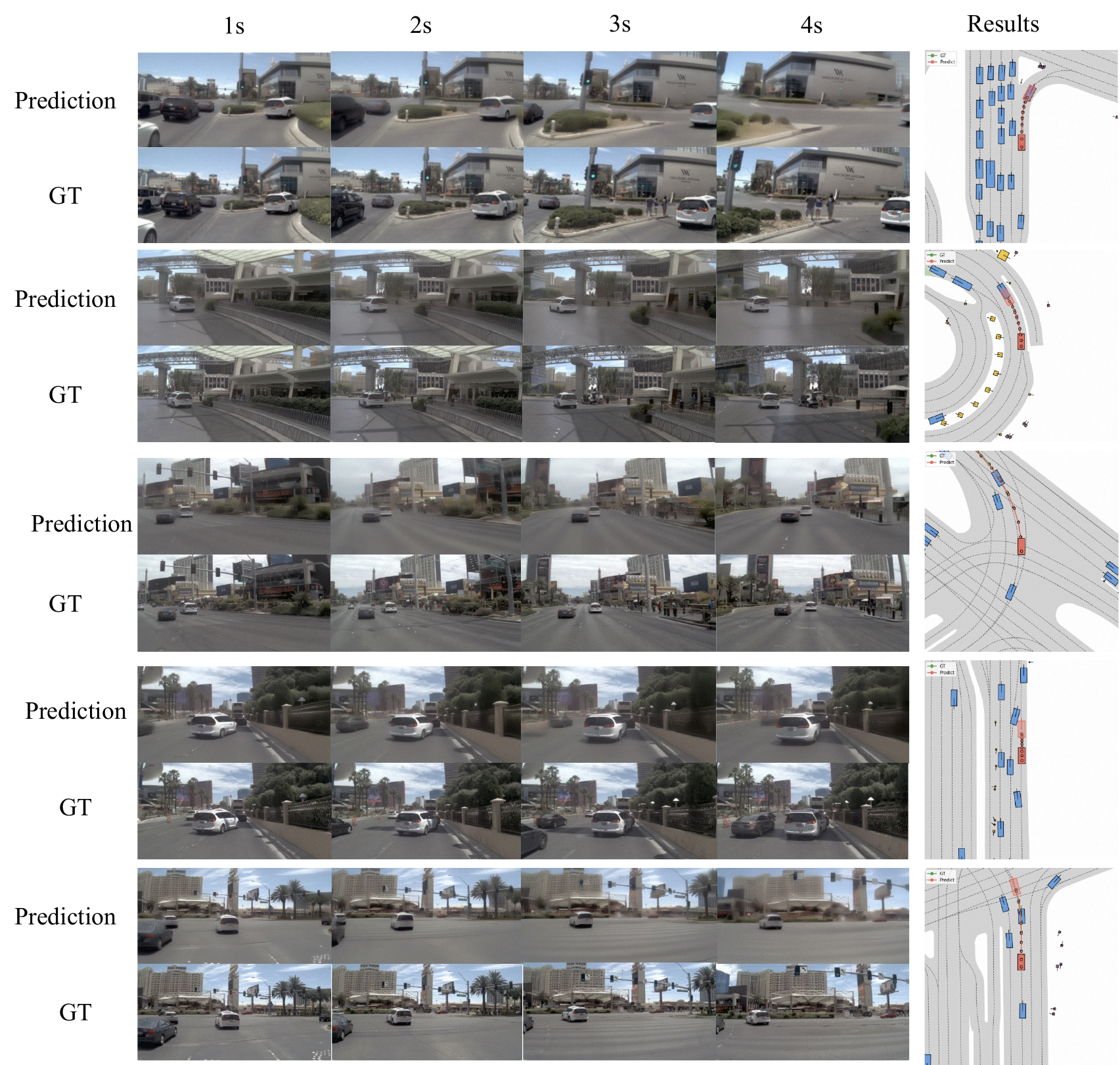} 
\caption{Visualization of planing and future prediction. The results are obtained using two denoising steps for both the image and the action.}
\label{fig5_vis_plan}
\end{figure*}

\begin{figure*}[t]
\centering
\includegraphics[width=0.9\textwidth]{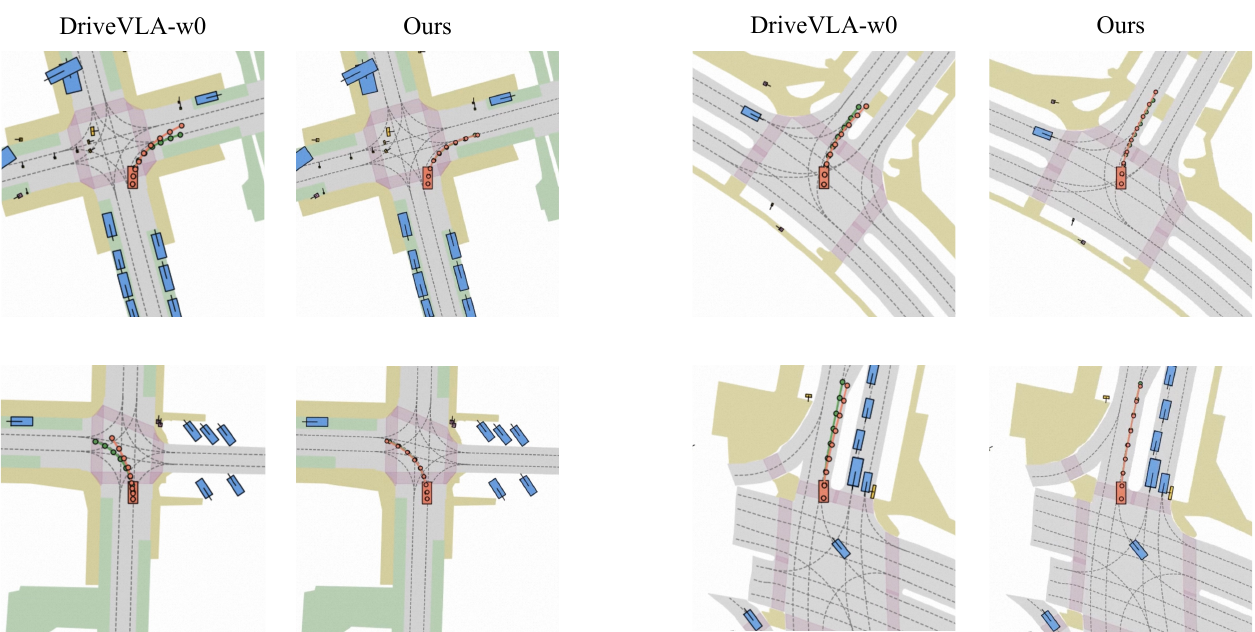} 
\caption{Comparison between our method and DriveVLA‑w0\cite{li2025drivevla}}
\label{fig6_vis_comp}
\end{figure*}

\paragraph{Visualization of Planning.} 
Figure~\ref{fig5_vis_plan} illustrates the performance of our method in several scenarios.
Our method not only produces accurate future scene predictions but also generates high-quality trajectory planning results. Figure~\ref{fig6_vis_comp} compares our method with the existing approach DriveVLA‑w0 \cite{li2025drivevla}, where it can be observed that our method exhibits better robustness in challenging situations such as turning scenarios.

\begin{figure*}[t]
\centering
\includegraphics[width=0.9\textwidth]{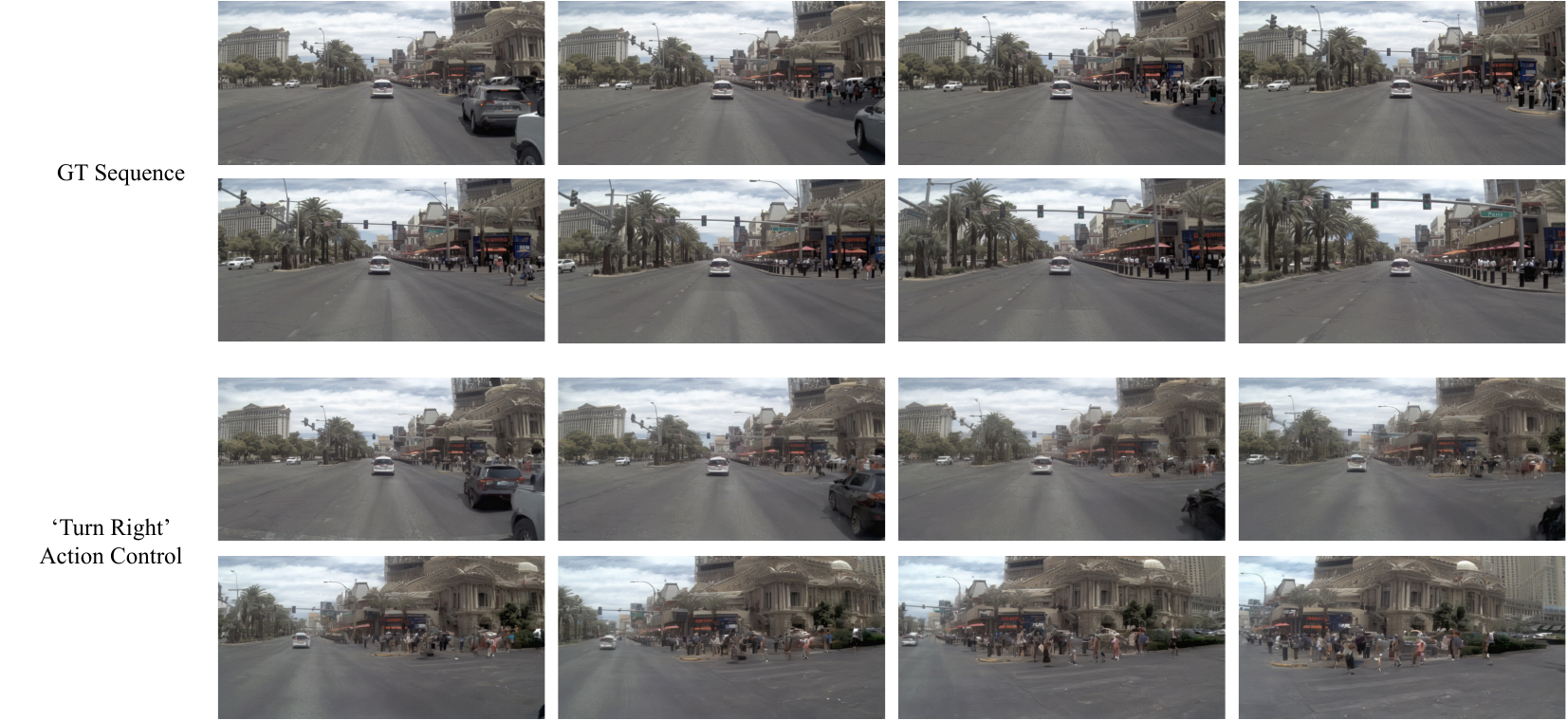} 
\caption{Visualization of action-Controlled video generation. The top half shows the ground-truth data, while the bottom half shows the generated video conditioned on a right‑turn trajectory.}
\label{fig7_vis_action}
\end{figure*}

\paragraph{Visualization of Action-Controlled Video Generation.}
Figure~\ref{fig7_vis_action} demonstrates our model's video generation capability under action control. The ground-truth data is a video of straight driving; our model, given the same conditional frames and a right-turn trajectory as input, can generate the corresponding video for a right-turn scenario.

\begin{figure*}[t]
\centering
\includegraphics[width=0.9\textwidth]{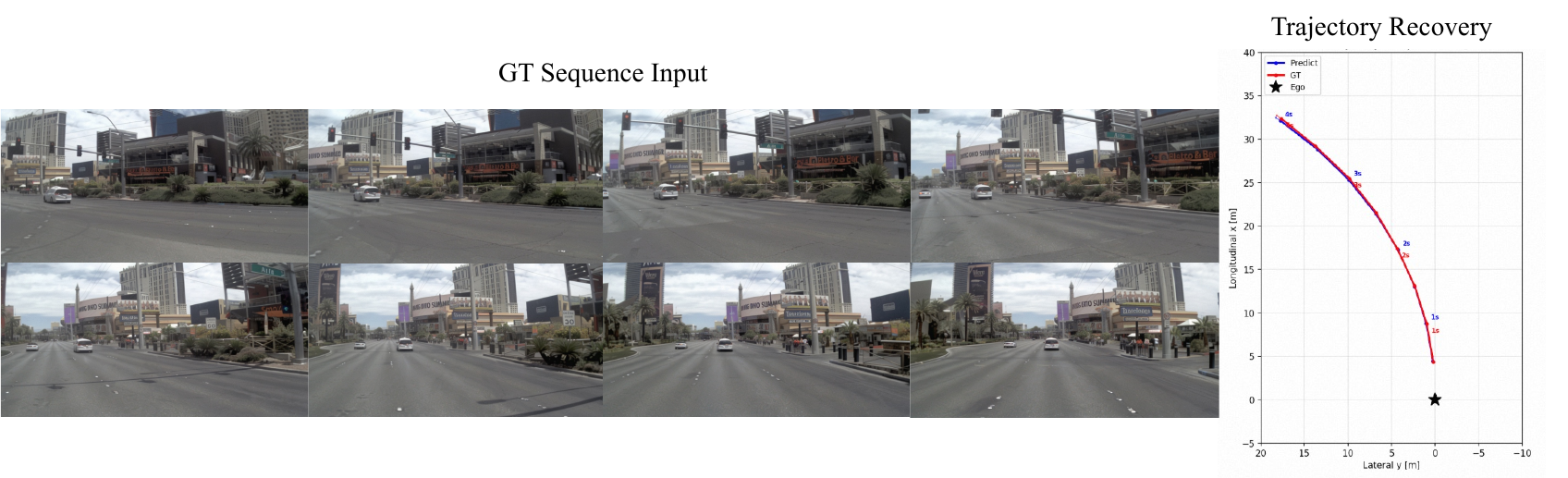} 
\caption{Visualization of trajectory recovery.}
\label{fig8_vis_idm}
\end{figure*}

\paragraph{Visualization of Trajectory Recovery.}
Figure~\ref{fig8_vis_idm} demonstrates the capability of our model as a visual odometry. Given a video of a turning scenario with large‑scale motion, our model can accurately recover the relative pose transformation between consecutive frames.

\begin{figure*}[t]
\centering
\includegraphics[width=0.9\textwidth]{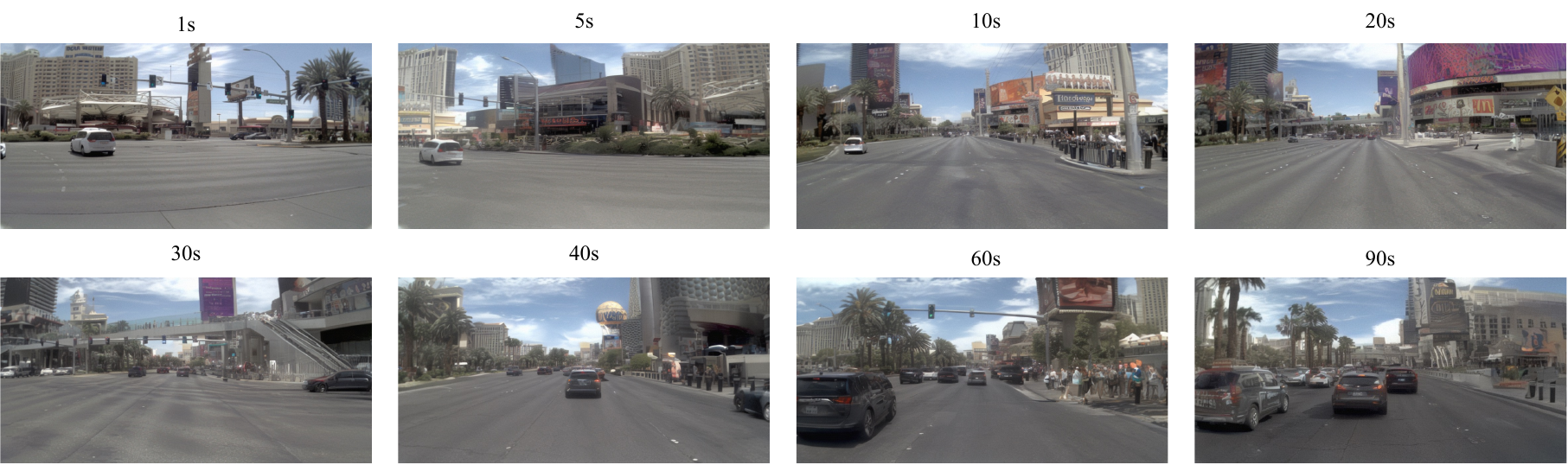} 
\caption{Visualization of long video generation.}
\label{fig9_vis_long}
\end{figure*}

\begin{figure*}[t]
\centering
\includegraphics[width=0.9\textwidth]{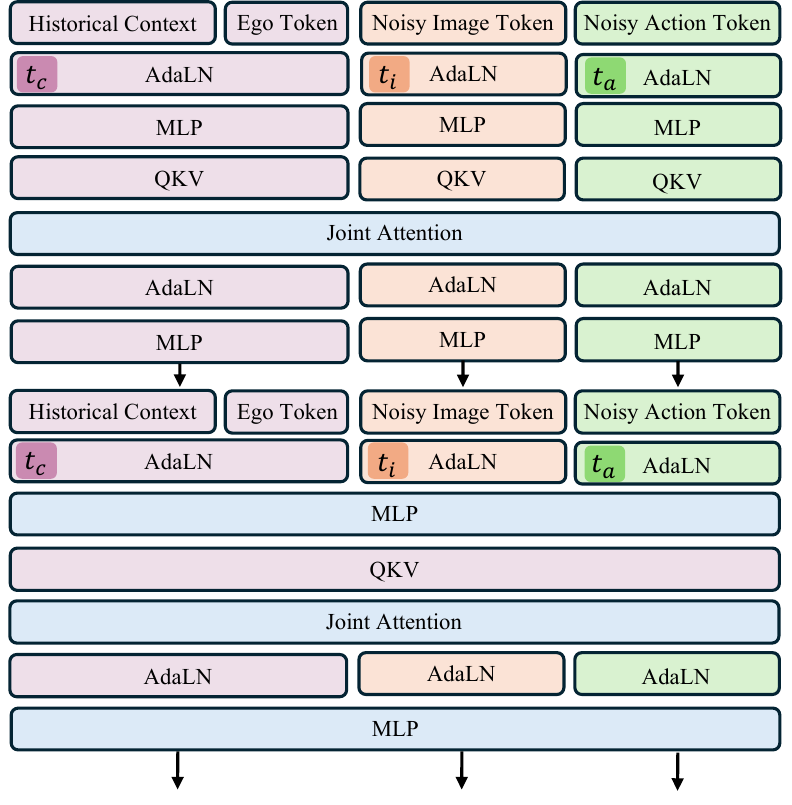} 
\caption{Detailed framework of UniDit. Our architecture is adapted from the existing video generation model \cite{zhang2025epona, labs2025flux}, and better inherits its capability of predicting future states.}
\label{fig10_unidit}
\end{figure*}

\paragraph{Visualization of Long Video Generation.}
Figure~\ref{fig9_vis_long} demonstrates our model's ability to generate long videos. Unlike some existing autonomous driving world action models that can only produce a fixed‑length video, our model supports the generation of videos with varying lengths. Moreover, when generating longer videos, it does not suffer from blurring and preserves high‑quality details of vehicles, roads, and buildings.

\begin{table*}[t]
    \centering
    \setlength{\tabcolsep}{4pt}
    \begin{tabular*}{\linewidth}{@{\extracolsep{\fill}}lcccccccccc}
        \toprule
        \textbf{Sampling steps} 
            & \textbf{NC}$\uparrow$ & \textbf{DAC}$\uparrow$ & \textbf{DDC}$\uparrow$ & \textbf{TL}$\uparrow$ & \textbf{EP}$\uparrow$ 
            & \textbf{TTC}$\uparrow$ & \textbf{LK}$\uparrow$ & \textbf{HC}$\uparrow$ & \textbf{EC}$\uparrow$ &  \textbf{EPDMS}$\uparrow$ \\
        \midrule
        2 & 98.8 & 97.5 & \textbf{99.7} & \textbf{99.8} & 87.3 & 98.3 & 98.0 & 98.3 & 85.7 & 89.9 \\
        4 & 98.9 & \textbf{97.6} & 99.6 & \textbf{99.8} & 87.2 & 98.4 & \textbf{98.1} & \textbf{98.4} & \textbf{86.5}  & \textbf{90.3} \\
        10 & \textbf{99.0} & 97.5 & 99.6 & \textbf{99.8} & 87.3 & \textbf{98.5} & \textbf{98.1} & \textbf{98.4} & 86.1 & 90.2 \\
        20 & 98.9 & 97.5 & 99.6 & \textbf{99.8} & \textbf{87.4} & 98.4 & 98.0 & \textbf{98.4} & 85.9  & 90.1 \\
        \bottomrule
    \end{tabular*}
    \caption{Ablation study on inference sampling steps. All results are obtained under the cascaded Act-then-Imagine inference paradigm, where the total sampling steps are evenly split between trajectory denoising and image denoising (e.g., 4 steps = 2 steps for trajectory + 2 steps for image).}
    \label{tab:sample}
\end{table*}

\begin{table*}[t]
    \centering
    \setlength{\tabcolsep}{4pt}
    \begin{tabular*}{\linewidth}{@{\extracolsep{\fill}}lcccccccccc}
        \toprule
        \textbf{Resolution} 
            & \textbf{NC}$\uparrow$ & \textbf{DAC}$\uparrow$ & \textbf{DDC}$\uparrow$ & \textbf{TL}$\uparrow$ & \textbf{EP}$\uparrow$ 
            & \textbf{TTC}$\uparrow$ & \textbf{LK}$\uparrow$ & \textbf{HC}$\uparrow$ & \textbf{EC}$\uparrow$ &  \textbf{EPDMS}$\uparrow$ \\
        \midrule
        256*512 & \textbf{99.0} & 97.2 & \textbf{99.6} & \textbf{99.9} & 86.6 & \textbf{98.5} & 98.0 & \textbf{98.3} & 81.7 & 89.3 \\
        512*1024 & 98.9 & \textbf{97.6} & \textbf{99.6} & 99.8 & \textbf{87.3} & 98.4 & \textbf{98.2} & \textbf{98.3} & \textbf{84.2}  & \textbf{89.9} \\
        \bottomrule
    \end{tabular*}
    \caption{Ablation study on input image resolution. All results are obtained under the Act-then-Imagine setting without ego-status co-training.}
    \label{tab:resolution}
\end{table*}

\begin{table*}[t]
\centering
\setlength{\tabcolsep}{4pt}{
\begin{tabular}{lccccccccccc}
\toprule
Method & Stage & NC & DAC & DDC & TLC & EP & TTC & LK & HC & EC & EPDMS \\
\midrule
\multirow{2}{*}{DiffusionDrive} & 1 & 96.8 & 86.0 & 98.8 & 99.3 & 84.0 & 95.8 & 96.7 & 97.6 & \textbf{79.6} & \multirow{2}{*}{27.5} \\
 & 2 & 80.1 & 72.8 & 84.4 & 98.4 & \textbf{85.9} & 76.6 & 46.4 & 96.3 & \textbf{72.8} & \\
\midrule
\multirow{2}{*}{MindDrive} & 1 & 96.1 & 86.0 & 98.8 & 99.3 & 83.3 & 95.6 & 94.4 & 97.6 & 74.7 & \multirow{2}{*}{30.5} \\
 & 2 & \textbf{82.6} & \textbf{79.1} & \textbf{86.4} & 98.0 & 85.3 & \textbf{79.4} & 49.2 & 96.5 & 71.0 & \\
\midrule
\multirow{2}{*}{DriveLaW} & 1 & 97.3 & 89.1 & \textbf{99.2} & \textbf{99.6} & \textbf{84.3} & \textbf{97.1} & 96.2 & \textbf{97.8} & 67.6 & \multirow{2}{*}{30.6} \\
 & 2 & 82.5 & 67.6 & 83.5 & 98.1 & 84.8 & 78.5 & 45.8 & 96.4 & 57.3 & \\
\midrule
\multirow{2}{*}{Ours} & 1 & \textbf{97.8} & \textbf{94.9} & 99.1 & 99.3 & 83.7 & 95.8 & \textbf{96.9} & \textbf{97.8} & 77.8 & \multirow{2}{*}{\textbf{34.8}} \\
 & 2 & 81.5 & 70.9 & 84.7 & \textbf{98.7} & 84.2 & 77.5 & \textbf{49.9} & \textbf{96.6} & 68.6 \\
\bottomrule
\end{tabular}}
\caption{Comparison on the NAVSIM v2 (navhard split) benchmark.}
\label{tab:navhard-v2}
\end{table*}

\begin{table*}[t]
\centering
\setlength{\tabcolsep}{4pt}{
\begin{tabular}{lccccccccccc}
\toprule
Method & Stage & NC & DAC & DDC & TLC & EP & TTC & LK & HC & EC & EPDMS \\
\midrule
\multirow{2}{*}{Imagine-then-Act} & 1 & \textbf{98.0} & 92.9 & \textbf{99.1} & \textbf{99.3} & \textbf{83.9} & \textbf{96.2} & \textbf{96.9} & \textbf{97.8} & 76.0 & \multirow{2}{*}{32.4} \\
 & 2 & 79.9 & 70.2 & 83.7 & 98.6 & \textbf{85.8} & 76.0 & 49.1 & 96.2 & 65.6 & \\
\midrule
\multirow{2}{*}{Imagine-and-Act} & 1 & 97.6 & 94.0 & \textbf{99.1} & 99.8 & 83.5 & 95.5 & \textbf{96.9} & \textbf{97.8} & \textbf{77.8} & \multirow{2}{*}{33.9} \\
 & 2 & 80.9 & 70.8 & 84.3 & 98.5 & 84.8 & 76.6 & 49.2 & 96.3 & \textbf{69.2} & \\
\midrule
\multirow{2}{*}{Act-then-Imagine} & 1 & 97.8 & \textbf{94.9} & \textbf{99.1} & \textbf{99.3} & 83.7 & 95.8 & \textbf{96.9} & \textbf{97.8} & \textbf{77.8} & \multirow{2}{*}{\textbf{34.8}} \\
 & 2 & \textbf{81.5} & \textbf{70.9} & \textbf{84.7} & \textbf{98.7} & 84.2 & \textbf{77.5} & \textbf{49.9} & \textbf{96.6} & 68.6 \\
\bottomrule
\end{tabular}}
\caption{Detailed results on the NAVSIM v2 (navhard split) benchmark with different inference mode.}
\label{tab:navhard_ablation}
\end{table*}

\end{document}